\pdfoutput=1

\documentclass[11pt,table,usenames,dvipsnames]{article}

\usepackage{acl}

\usepackage{times}
\usepackage{latexsym}
\usepackage{hyperref}
\usepackage{tabularx}
\usepackage{color}
\usepackage{xcolor}
\usepackage{graphicx}
\usepackage{multirow}
\usepackage{float}
\usepackage{caption}

\usepackage[T1]{fontenc}

\usepackage[utf8]{inputenc}

\usepackage{microtype}

\definecolor{bloomcolor}{HTML}{a1c9f4}
\definecolor{falconcolor}{HTML}{ffb482}
\definecolor{llamacolor}{HTML}{8de5a1}
\definecolor{mistralcolor}{HTML}{ff9f9b}
\definecolor{mptcolor}{HTML}{d0bbff}
\definecolor{phicolor}{HTML}{debb9b}

\title{From Test-Taking to Test-Making: Examining LLM Authoring of Commonsense Assessment Items}

\author{Melissa Roemmele \\
  Midjourney \\
  San Francisco, CA, USA\\
  \texttt{mroemmele@midjourney.com} \\\And
  Andrew S. Gordon \\
  Institute for Creative Technologies\\
  University of Southern California\\
  Los Angeles, CA, USA \\
  \texttt{gordon@ict.usc.edu} \\}

\begin{document}
\maketitle
\begin{abstract}
LLMs can now perform a variety of complex writing tasks. They also excel in answering questions pertaining to natural language inference and commonsense reasoning. Composing these questions is itself a skilled writing task, so in this paper we consider LLMs as authors of commonsense assessment items. We prompt LLMs to generate items in the style of a prominent benchmark for commonsense reasoning, the Choice of Plausible Alternatives (COPA). We examine the outcome according to analyses facilitated by the LLMs and human annotation. We find that LLMs that succeed in answering the original COPA benchmark are also more successful in authoring their own items.
\end{abstract}

\section{Introduction}

Large Language Models (LLMs) can perform complex writing tasks ranging from paraphrasing sentences to composing long-form stories. Because success on these open-ended authoring tasks is hard to measure quantitatively, NLP researchers tend to focus on more constrained tasks when judging the abilities of LLMs. Many of these assessment tasks, or \textit{benchmarks}, are conceptually connected with capabilities relevant to authoring, but they use a specific data design to support quantitative evaluation. For instance, assessments of commonsense reasoning often involve multiple-choice question answering, and accuracy on these questions is a proxy indicator for a model's ability to write coherently. New LLMs have excelled on these types of evaluations, but how this success relates to the LLMs' authoring capabilities is still unclear.

Creating these assessments is itself a skilled writing task, which up to this point has been performed by human authors. Given their success in answering assessments, in this paper we examine whether LLMs can author assessment items as well. We prompt LLMs to generate items in the style of one well-known benchmark for commonsense reasoning, the Choice of Plausible Alternatives (COPA). We examine the outcome in terms of the LLMs' responses to their own generated items as well to what degree the items meet the benchmark authoring standards as judged by human raters. We show that an LLM's authoring success is associated with its success in answering the original benchmark.

\section{COPA}

The Choice of Plausible Alternatives (COPA) \citep{roemmele2011choice} is an English-language benchmark that assesses the task of commonsense causal reasoning. As shown by the examples in Table \ref{example_copa_items}, each item in COPA consists of a \textit{premise}, a question specifying a causal \textit{direction} (forwards or backwards), and two \textit{alternatives}. One alternative is considered more plausible than the other with regard to the question. For the forwards direction, the question elicits the alternative that is the more plausible effect (result) of the premise, whereas for the backwards direction the question elicits the more plausible cause of the premise. Human performance on this benchmark is considered 100\%. The order of the alternatives is balanced across the test set such that random guessing yields 50\% accuracy.

\begin{table}[h]
\small
\centering
\rowcolors{1}{white}{gray!20}
\begin{tabularx}{\columnwidth}{ X }
\hline
\textbf{Premise:} The girl received a trophy.\newline
What was the \textit{cause} of this?\newline
\textbf{Alternative 1:} She won a spelling bee.\newline
\textbf{Alternative 2:} She made a new friend.\\

\textbf{Premise:} I tipped the bottle.\newline
What happened as a \textit{result}?\newline
\textbf{Alternative 1:} The liquid in the bottle froze.\newline
\textbf{Alternative 2:} The liquid in the bottle poured out.
\\
\hline
\end{tabularx}
\caption{Examples of COPA items}
\label{example_copa_items}
\end{table}

From an authoring perspective, creating a COPA item involves writing a collection of one-sentence narrations of events. The author writes two events that have a clear cause-and-effect relation, which respectively become the premise and the more plausible alternative (\texttt{mpa}). The author also writes an event that does not have a clear causal relation to the premise, and this becomes the less plausible alternative (\texttt{lpa}). As described in \citet{roemmele2011choice}, composing the \texttt{lpa} is particularly challenging. This is because the benchmark is intended to assess models' ability to isolate causal relations in text separate from more generic associations that are captured by simple lexical co-occurence statistics. The author is expected to write an \texttt{lpa} that has some semantic and/or temporal relation to the premise, but no salient causal relation.



Until only recently, COPA was considered a difficult benchmark. When first presented as a shared task at SemEval-2012, submitted systems achieved accuracy in the 60-65\% range \citep{gordon-etal-2012-semeval}. When COPA was incorporated into the well-known SuperGLUE ensemble of benchmarks in 2019, the best system reached almost 85\% \citep{wang2019b}. Today some LLMs obtain near-perfect accuracy, as we observe in this work. This reflects the life cycle of most benchmarks, where existing ones are continuously replaced by harder new ones once maximal performance is obtained.

\section{Research Questions}

In this work, we apply a common LLM interaction paradigm (prompting with exemplars) to facilitate LLMs to generate COPA-style items. Moving forward in this paper, we refer to the original COPA items as \textbf{Orig-COPA} and the LLM-authored items as \textbf{Gen-COPA}. When existing research refers to the COPA task, it means the task of predicting the answer (i.e. more plausible alternative) to each item. Here, we more specifically refer to this task as \textbf{answering} COPA, in order to distinguish it from the task of generating COPA items. 

We are interested in the following research questions. First, \textbf{can LLMs author items with the design of COPA?} In particular, how often do Gen-COPA items meet the same authoring standards as the original benchmark? Second, \textbf{when an LLM produces its own Gen-COPA set, does it then answer its own items correctly?} In other words, does the LLM behave consistently during generation and answering in what it deems as the more plausible alternative? Third, \textbf{how does an LLM’s success in authoring Gen-COPA items relate to its ability to correctly answer Orig-COPA items?} Are LLMs that do well on Orig-COPA also better at authoring new COPA items? 


\section{Selected LLMs}\label{llms}

\begin{table*}[h!]
\small
\centering
\begin{tabularx}{\textwidth}{ X  p{.4175\textwidth} }
\textbf{Prompt Template} & \textbf{Prompt Example}\\
\hline
\scriptsize\texttt{\{\% for ex in exemplars \%\}\newline
Premise: \{\{ex[`premise']\}\}\newline
\{\% if ex[`direction'] == `backwards' \%\}What was the cause of this?\{\% else \%\}What happened as a result?\{\% endif \%\}\newline
Alternative 1: \{\{ex[`alt1']\}\}\newline
Alternative 2: \{\{ex[`alt2']\}\}\newline
The more plausible \{\% if ex[`direction'] == `backwards' \%\}cause\{\% else \%\}result\{\% endif \%\} is Alternative \{\{ex[`more\_plausible\_alt']\}\}.\newline
\{\% endfor \%\}\newline\newline
Premise: \{\{item[`premise']\}\}\newline
\{\% if item[`backwards'] == `cause' \%\}What was the cause of this?\{\% else \%\}What happened as a result?\{\% endif \%\}\newline
Alternative 1: \{\{item[`alt1']\}\}\newline
Alternative 2: \{\{item[`alt2']\}\}\newline
The more plausible \{\% if item[`backwards'] == `cause' \%\}cause\{\% else \%\}result\{\% endif \%\} is Alternative} & 
\cellcolor{gray!20}Premise: My body cast a shadow over the grass.\newline
What was the cause of this?\newline
Alternative 1: The sun was rising.\newline
Alternative 2: The grass was cut.\newline
The more plausible cause is Alternative 1.\newline
\newline
\textit{[...3 more exemplars...]}
\newline\newline
Premise: My favorite song came on the radio.\newline
What happened as a result?\newline
Alternative 1: I covered my ears.\newline
Alternative 2: I sang along to it.\newline
The more plausible result is Alternative\\
\hline
\end{tabularx}
\caption{4-shot prompt design for answering COPA items. The left shows the raw template (in Jinja2 syntax). The rendered version provided as input to the LLM is on the right.}
\label{copa_answering_prompt}
\end{table*}

To investigate these questions, we employ a variety of notable open-source LLMs. We consider eleven models from six different families, where models within the same family are distinguished by size (number of parameters).

\colorbox{bloomcolor}{\textsc{bloom-7b}} \& \colorbox{bloomcolor}{\textsc{bloom-176b}}\footnote{huggingface.co/bigscience/\href{https://huggingface.co/bigscience/bloom}{\{bloom},\href{https://huggingface.co/bigscience/bloom-7b1}{bloom-7b1\}}}: The BigScience Large Open-science Open-access Multilingual Language Model (\textsc{bloom}) family was developed through a large cross-team research collaboration \citep{workshop2022bloom}. It was trained on a collection of publicly available datasets that together comprise 1.61 terabytes of text spanning a wide variety of natural and programming languages \citep{laurencon2022the}. Its performance on various benchmarks in SuperGLUE was competitive with other open-source models.

\colorbox{falconcolor}{\textsc{falcon-7b}} \& \colorbox{falconcolor}{\textsc{falcon-40b}}\footnote{huggingface.co/tiiuae/\href{https://huggingface.co/tiiuae/falcon-7b}{\{falcon-7b},\href{https://huggingface.co/tiiuae/falcon-40b}{falcon-40b\}}}: The \textsc{falcon} family \citep{almazrouei2023falconseriesopenlanguage} developed by the Technology Innovation Institute was trained on 1-1.5 trillion tokens, primarily the RefinedWeb corpus derived from CommonCrawl \citep{refinedweb}. The 40B-sized model obtained top performance on several benchmarks upon its release.

\colorbox{llamacolor}{\textsc{llama2-7b}}, \colorbox{llamacolor}{\textsc{llama2-13b}}, \& \colorbox{llamacolor}{\textsc{llama2-70b}}\footnote{huggingface.co/meta-llama/\href{https://huggingface.co/meta-llama/Llama-2-7b-hf}{\{Llama-2-7b},\href{https://huggingface.co/meta-llama/Llama-2-13b-hf}{Llama-2-13b},\href{https://huggingface.co/meta-llama/Llama-2-70b-hf}{Llama-2-70b\}}}: The \textsc{Llama2} family developed by Meta  was trained on a collection of publicly available datasets comprising 2 million tokens \citep{touvron2023llama}. It has outperformed \textsc{falcon} and \textsc{mpt} (below) on several benchmarks.

\colorbox{mistralcolor}{\textsc{mistral-7b}}\footnote{\href{https://huggingface.co/mistralai/Mistral-7B-v0.3}{huggingface.co/mistralai/Mistral-7B-v0.3}} \textsc{mistral-7b} \citep{jiang2023mistral}, developed by Mistral AI, was trained on open web data. Upon its release, it was presented as a key competitor of the \textsc{llama2} models. It outperformed the much larger \textsc{llama2-13b} on various knowledge and reasoning benchmarks.

\colorbox{mptcolor}{\textsc{mpt-7b}} \& \colorbox{mptcolor}{\textsc{mpt-30b}}\footnote{huggingface.co/mosaicml/\href{https://huggingface.co/mosaicml/mpt-7b}{\{mpt-7b},\href{https://huggingface.co/mosaicml/mpt-30b}{mpt-30b\}}}: The MosaicML Pretrained Transformer (\textsc{MPT}) family developed by MosaicML was trained on 1 billion tokens from a mix of web text and curated data \citep{MosaicML2023Introducing}. It has performed favorably compared with other open-source LLMs on benchmarks including COPA.

\colorbox{phicolor}{\textsc{phi-2}}\footnote{\href{https://huggingface.co/microsoft/phi-2}{huggingface.co/microsoft/phi-2}}: \textsc{phi-2} \citep{Javaheripi2023}, developed by Microsoft, is a 2.7B parameter model trained on 250 billion tokens from ``textbook quality'' data \citep{gunasekar2023textbooks}. Some texts were selectively curated from the web while others were synthetically generated, with both processes emphasizing a high density of commonsense knowledge in the resulting data. Upon its release, \textsc{phi-2} demonstrated similar or better performance on commonsense reasoning benchmarks compared with the much larger \textsc{mistral} and \textsc{llama2} models.

\section{Orig-COPA Answering Performance}\label{original_copa}

We first consider how well these LLMs answer the original COPA benchmark. We used a few-shot prompting approach to elicit answers for COPA items. In particular, we selected four items from the COPA development set as task exemplars to include in all prompts for the 500 items in the test set.  Table \ref{copa_answering_prompt} shows the prompt format with the exemplars preceding the target item to be answered\footnote{All prompt templates and code for reproducing this work are available at: \href{https://github.com/roemmele/Gen-COPA}{github.com/roemmele/Gen-COPA}}. Each exemplar ends with a sentence specifying the numerical label of the correct (more plausible) alternative, and the target item includes the prefix of this sentence in order to elicit a corresponding prediction from the model. 

We applied the LLMs\footnote{We ran all LLMs using HuggingFace inference endpoints: \href{https://huggingface.co/inference-endpoints}{huggingface.co/inference-endpoints}} outlined in Section \ref{llms} to these prompts for the COPA test set. We used greedy decoding (i.e. selecting the maximum probability token at each generation step) to ensure the predictions were deterministic. We generated outputs with a maximum length of four tokens, then post-processed each output with a regular expression to detect the presence of ``1'' or ``2'' as the predicted answer. Our policy was to randomly select one of these numbers if neither was present in the output, but in our experiments this was not applied since all outputs contained a 1 or 2.

Table \ref{copa_answering_accuracy} shows the accuracy of these predictions according to the benchmark. For families with models of varying sizes, the larger models outperform the smaller ones within each family, which is an expected pattern \citep[e.g.][]{kaplan2020scaling}. However, size does not fully account for the overall ranking of model performance. For example, \textsc{bloom-176b} is the largest model but ranks only the third lowest in accuracy, and \textsc{phi-2} is the smallest model but ranks the third highest. 

\begin{table}[h]
\small
\centering
\begin{tabular}{ c  c }
\textbf{Model} & \textbf{Accuracy}\\
\hline
\rowcolor{bloomcolor}\textsc{bloom-7b} & 0.532\\
\rowcolor{bloomcolor}\textsc{bloom-176b} & 0.576\\
\rowcolor{falconcolor}\textsc{falcon-7b} & 0.538\\
\rowcolor{falconcolor}\textsc{falcon-40b} & 0.844\\
\rowcolor{llamacolor}\textsc{llama2-7b} & 0.826\\
\rowcolor{llamacolor}\textsc{llama2-13b} & 0.850\\
\rowcolor{llamacolor}\textsc{llama2-70b} & 0.976\\
\rowcolor{mistralcolor}\textsc{mistral-7b} & 0.938\\
\rowcolor{mptcolor}\textsc{mpt-7b} & 0.656\\
\rowcolor{mptcolor}\textsc{mpt-30b} & 0.848\\
\rowcolor{phicolor}\textsc{phi-2} & 0.902\\
\hline
\end{tabular}
\caption{Answering accuracy on COPA test set}
\label{copa_answering_accuracy}
\end{table}

\begin{table*}[h!]
\small
\centering
\begin{tabularx}{\textwidth}{ X  p{.4\textwidth} }
\textbf{Prompt Template} & \textbf{Prompt Example}\\
\hline
\scriptsize\texttt{\{\% for ex in exemplars \%\}\newline
Premise: \{\{ex[`premise']\}\}\newline
\{\% if ex[`direction'] == `backwards' \%\}What was the cause of this?\{\% else \%\}What happened as a result?\{\% endif \%\}\newline
More Plausible Alternative: \{\% if ex[`more\_plausible\_alt'] == `1' \%\}\{\{ex[`alt1']\}\}\{\% else \%\}\{\{ex[`alt2']\}\}\{\% endif \%\}\newline
Less Plausible Alternative: \{\% if ex[`more\_plausible\_alt'] == `1' \%\}\{\{ex[`alt2']\}\}\{\% else \%\}\{\{ex[`alt1']\}\}\{\% endif \%\}\newline
\{\% endfor \%\}\newline
\newline
Premise:}
& \cellcolor{gray!20}Premise: The girl politely declined the hamburger.\newline
What was the cause of this?\newline
More Plausible Alternative: She was a vegetarian.\newline
Less Plausible Alternative: She liked fast food.\newline
\newline
\textit{[...2 more exemplars...]}
\newline\newline
Premise:\\
\hline
\end{tabularx}
\caption{3-shot prompt design for generating COPA items}
\label{copa_generating_prompt}
\end{table*}

Even with the wide spread in performance between different models, the near-perfect accuracy from \textsc{llama2-70b} confirms that COPA is no longer as challenging as it previously was and will soon become outdated. We note that because the COPA test set is available online, it is possible the LLMs observed these items during training. While we expect that developers of these LLMs adhered to the standard best practice of excluding test data from training sets, we cannot strictly assume this. This issue is further addressed in the next section.

\section{LLM Authoring of Gen-COPA}\label{llm_authoring_gen_copa}

Next, we examined to what degree these LLMs can generate their own items in the design of COPA. We again used a few-shot prompting approach to facilitate this task. Each prompt consisted of three items of one particular causal direction (forwards or backwards) randomly sampled from the COPA development set. Table \ref{copa_generating_prompt} shows the format: instead of using the labels ``Alternative 1'' and ``Alternative 2'' as in the answering task, each exemplar directly refers to the ``More Plausible Alternative'' and ``Less Plausible Alternative'' for a given premise and question, and the models are expected to generate segments with these same identifiers.

We assembled 500 prompts for each causal direction, so each LLM was run on 1000 prompts total. Each of these prompts had a unique set of exemplars. To further promote diversity in the outputs, we used random sampling during decoding, in particular top-p (nucleus) sampling with p=0.9 and temperature=1.0. We generated outputs with a maximum length of 200 tokens.

We parsed each LLM output with the template shown in Table \ref{gen_copa_parsing_template}, which captures the premise, more plausible alternative (\texttt{mpa}), and less plausible alternative (\texttt{lpa}) segments for each item of a pre-defined direction. We refer to these resulting variables for a single output as a \textbf{schema}. We automatically classified an output as \textit{failed} if it could not be parsed according to the template or if at least one segment in the item exactly matched another one (e.g. the two alternatives were the same). We also considered whether the generated outputs were duplicates of Orig-COPA items. We failed any output whose tokens were all contained in a single item from the Orig-COPA dev or test set. See Appendix \ref{gen_copa_novelty_section} for further validation of the novelty of the Gen-COPA items in relation to Orig-COPA. In Table \ref{gen_copa_consistency}, \textbf{\# Items} refers to the number of \textit{passable} (non-failed) schemas generated by each LLM. Across all LLMs the vast majority of outputs (between 97.6\% and 99.2\%) are passable.\footnote{A remark about failures resulting from duplicating (plagiarizing) Orig-COPA items: a further analysis of these outputs across all LLMs revealed that the plagiarized item was always one of the prompt exemplars. We did not encounter cases where an LLM  duplicated an Orig-COPA item that was not contained in the prompt.}

\begin{table}[h!]
\footnotesize
\centering
\begin{tabularx}{\linewidth}{| X |}
\hline
\texttt{<premise>\newline
\{\% if direction == `backwards' \%\}What was the cause of this?\{\% else \%\}What happened as a result?\{\% endif \%\}\newline
More Plausible Alternative: <mpa>\newline
Less Plausible Alternative: <lpa>}\\
\hline
\end{tabularx}
\caption{Parsing template for Gen-COPA LLM outputs}
\label{gen_copa_parsing_template}
\end{table}

\subsection{Consistency}\label{consistency}

Our first analysis of the resulting Gen-COPA items assessed the LLMs' \textbf{consistency} between generation and answering. In particular, when an LLM is presented with an item that the LLM itself generated, is the alternative it predicts as more plausible the same one it originally authored as the \texttt{mpa}? To determine this, we transformed each LLM's passable schemas into the design of the benchmark items by randomly assigning the labels of ``Alternative 1'' and ``Alternative 2'' to the \texttt{mpa} and \texttt{lpa} in each schema. We ensured each resulting Gen-COPA set was balanced such that always guessing Alternative 1 as the answer yielded 50\% accuracy (or trivially above 50\% for sets with an odd number of items). We used the same 4-shot prompt design we applied to Orig-COPA (Section \ref{original_copa}) in order to elicit answers from an LLM for its own Gen-COPA set. We quantify consistency as the LLM's answering accuracy on this set: it measures how often the LLM's predicted answer for an item is the same alternative that LLM originally designated as the \texttt{mpa} in the generated schema for that item. 

Table \ref{gen_copa_consistency} shows that consistency varies widely between different LLMs. The least consistent models (\textsc{bloom-7b}, \textsc{falcon-7b}) predict their \texttt{mpa} only slightly more often than random chance, while the most consistent model (\textsc{llama2-70b}) predicts it about 86\% of the time. Thus, LLMs are not guaranteed to perform well on their own generated items. Notably, consistency is associated with answering accuracy on Orig-COPA, which is indicated by the extremely strong correlation (in terms of Spearman's rank-order) between the scores in Table \ref{copa_answering_accuracy} and \ref{gen_copa_consistency} ($r_s=.97, p<.001$).

\begin{table}[h]
\small
\centering
\begin{tabular}{ c  c  c }
\textbf{Model} & \textbf{\# Items} & \textbf{Consistency}\\
\hline
\rowcolor{bloomcolor}\textsc{bloom-7b} & 983 & 0.512\\
\rowcolor{bloomcolor}\textsc{bloom-176b} & 980 & 0.588\\
\rowcolor{falconcolor}\textsc{falcon-7b} & 992 & 0.505\\
\rowcolor{falconcolor}\textsc{falcon-40b} & 987 & 0.733\\
\rowcolor{llamacolor}\textsc{llama2-7b} & 976 & 0.673\\
\rowcolor{llamacolor}\textsc{llama2-13b} & 986 & 0.753\\
\rowcolor{llamacolor}\textsc{llama2-70b} & 978 & 0.858\\
\rowcolor{mistralcolor}\textsc{mistral-7b} & 993 & 0.777\\
\rowcolor{mptcolor}\textsc{mpt-7b} & 989 & 0.602\\
\rowcolor{mptcolor}\textsc{mpt-30b} & 991 & 0.701\\
\rowcolor{phicolor}\textsc{phi-2} & 987 & 0.828\\
\hline
\end{tabular}
\caption{Answering accuracy of each LLM on its own Gen-COPA set (i.e. consistency)}
\label{gen_copa_consistency}
\end{table}

\subsection{Validity}\label{gen_copa_validity_section}

The consistency analysis does not indicate whether the Gen-COPA items are \textbf{valid} according to a standard other than the LLM itself. Here, we consider an item valid if the alternative deemed more plausible by a consensus of humans is the same as the \texttt{mpa} in the generated schema. This is analogous to the development of the original COPA benchmark, where an item was considered valid if two human judges both selected the answer designated as the \texttt{mpa} by the author of the item.

To assess the validity of the Gen-COPA items, we employed a two-stage process. In the first stage, an internally employed expert rater reviewed each schema to judge if the \texttt{mpa} was indeed more plausible than the \texttt{lpa}. If so, the rater marked the schema as conditionally-valid, otherwise they marked it as invalid. We then set aside the invalid schemas and converted the conditionally-valid schemas to items with the randomized ``Alternative 1'' and ``Alternative 2'' labels. We presented each of these items to two external raters on the Prolific\footnote{\href{https://www.prolific.com/}{prolific.com}} data collection platform. The raters observed the premise, question, and alternatives for each item and selected from two options indicating either Alternative 1 or 2 as the more plausible answer to the question. Ultimately, an item was considered fully-valid if both Prolific raters selected the alternative that was also designated as the \texttt{mpa} in the schema for that item. Items where raters disagreed on the answer were marked as invalid. Each rater responded to 50 items and was paid \$6 for an expected completion time of no more than 30 minutes.

We randomly sampled 300 schemas per LLM for this validity assessment, a total of 3300 items. The expert rater flagged a few items to be withheld from the external raters due to potentially offensive content. We resampled a new item to replace each of these (0-5 per LLM, as shown in Table \ref{gen_copa_validity}). 1033 items were classified as conditionally-valid in the first stage of annotation. In the second stage, the two raters agreed on the \texttt{mpa} for 914 of them. Their agreement in terms of Cohen’s $\kappa$ was .79, indicating substantial agreement. 

Table \ref{gen_copa_validity} shows the proportion of items for each LLM that were categorized as valid through this process. The validity rate varies significantly between models. The least successful LLM is \textsc{bloom-7b}, with only 10\% of items marked valid, while \textsc{llama2-70b} has the most success with $\approx$46\% of its items marked valid. Just like consistency, validity is strongly correlated with answering accuracy on Orig-COPA ($r_s=.87, p<.001$). Thus, models that perform well on the original benchmark are more likely to generate valid items.

\begin{table}[h]
\small
\centering
\begin{tabular}{ c  c  c }
\textbf{Model} & \textbf{\# Replaced} & \textbf{Validity}\\
\hline
\rowcolor{bloomcolor}\textsc{bloom-7b} & 3 & 0.100\\
\rowcolor{bloomcolor}\textsc{bloom-176b} & 2 & 0.227\\
\rowcolor{falconcolor}\textsc{falcon-7b} & 0 & 0.113\\
\rowcolor{falconcolor}\textsc{falcon-40b} & 4 & 0.343\\
\rowcolor{llamacolor}\textsc{llama2-7b} & 3 & 0.197\\
\rowcolor{llamacolor}\textsc{llama2-13b} & 1 & 0.330\\
\rowcolor{llamacolor}\textsc{llama2-70b} & 2 & 0.463\\
\rowcolor{mistralcolor}\textsc{mistral-7b} & 3 & 0.310\\
\rowcolor{mptcolor}\textsc{mpt-7b} & 5 & 0.223\\
\rowcolor{mptcolor}\textsc{mpt-30b} & 2 & 0.300\\
\rowcolor{phicolor}\textsc{phi-2} & 1 & 0.440\\
\hline
\end{tabular}
\caption{Validity rate of Gen-COPA items}
\label{gen_copa_validity}
\end{table}

We conducted a qualitative analysis of the invalid items to characterize the most common reasons they were rejected. Table \ref{invalid_patterns} lists these characteristics with some exemplifying schemas. 

\begin{table*}[h!]
\small
\centering
\rowcolors{1}{white}{gray!20}
\begin{tabularx}{\textwidth}{ p{0.28\textwidth}  X }
\textbf{Description} & \textbf{Example}\\
\hline
The premise is vague or difficult to interpret  &
\textbf{Premise}: It happened on a blacktop road. 
What happened as a \textit{result}? [\textsc{bloom-7b}]\newline
\textbf{\texttt{mpa}}: The car slid on the ice.\newline
\textbf{\texttt{lpa}}: The car jumped into the ditch.\\

The \texttt{mpa} is not plausible in relation to the premise &
\textbf{Premise}: The woman contracted polio.
What was the \textit{cause} of this? [\textsc{bloom-7b}]\newline
\textbf{\texttt{mpa}}: She ate ice cream.\newline
\textbf{\texttt{lpa}}: She ran into a car.\\

The \texttt{mpa} has a semantic or temporal relation to the premise, but not a clear \textit{causal} relation that distinguishes it from the \texttt{lpa} &
\textbf{Premise}: We opened the envelope.
What happened as a \textit{result}? [\textsc{llama2-7b}]\newline
\textbf{\texttt{mpa}}: We found \$1,000 inside.\newline
\textbf{\texttt{lpa}}: We found nothing.\\

The \texttt{mpa} is not anchored to commonsense knowledge &
\textbf{Premise}: A balloon burst. 
What was the \textit{cause} of this? [\textsc{bloom-7b}]\newline
\textbf{\texttt{mpa}}: The balloon was too big for its zipper.\newline
\textbf{\texttt{lpa}}: The balloon was too small for its zipper.\\

Some causal information in the \texttt{mpa} is already contained in the premise &
\textbf{Premise}: She was fired for showing up late.
What was the \textit{cause} of this? [\textsc{mpt-30b}]\newline
\textbf{\texttt{mpa}}: She arrived late for work.\newline
\textbf{\texttt{lpa}}: She arrived too early for work.\\


The \texttt{mpa} and \texttt{lpa} are synonymous or closely related in meaning &
\textbf{Premise}: The man stubbed his toe. 
What happened as a \textit{result}? [\textsc{falcon-40b}]\newline
\textbf{\texttt{mpa}}: He flinched.\newline
\textbf{\texttt{lpa}}: He winced.\\

The \texttt{mpa} and \texttt{lpa} are both plausible to an equal degree &
\textbf{Premise}: I ate the moldy bread. 
What happened as a \textit{result}? [\textsc{falcon-7b}]\newline
\textbf{\texttt{mpa}}: I vomited.\newline
\textbf{\texttt{lpa}}: I developed an unusual stomachache.\\

Assessing the relative plausibility of the \texttt{mpa} and \texttt{lpa} is subjective or requires more information &
\textbf{Premise}: The boy wanted to be wealthy. 
What happened as a \textit{result}? [\textsc{llama2-7b}]\newline
\textbf{\texttt{mpa}}: He started a company.\newline
\textbf{\texttt{lpa}}: He stole the money of others.\\
\hline
\end{tabularx}
\caption{Common characteristics of invalid Gen-COPA items}
\label{invalid_patterns}
\end{table*}

\subsection{Gen-COPA Answering Performance}\label{gen_copa_answering_performance}

\begin{table*}[t]
\small
\centering
\begin{tabular}{| c | c | c | c | c | c | c | c | c | c | c | c | c |}
\hline
\textbf{Item Set} $\rightarrow$ & \rotatebox[origin=c]{90}{\textsc{bloom-7b}} & 
\rotatebox[origin=c]{90}{\textsc{bloom-176b}} &
\rotatebox[origin=c]{90}{\textsc{falcon-7b}} &
\rotatebox[origin=c]{90}{\textsc{falcon-40b}} &
\rotatebox[origin=c]{90}{\textsc{llama2-7b}} &
\rotatebox[origin=c]{90}{\textsc{llama2-13b}} &
\rotatebox[origin=c]{90}{\textsc{llama2-70b}} &
\rotatebox[origin=c]{90}{\textsc{mistral-7b}} &
\rotatebox[origin=c]{90}{\textsc{mpt-7b}} &
\rotatebox[origin=c]{90}{\textsc{mpt-30b}} &
\rotatebox[origin=c]{90}{\textsc{phi-2}} &
\rotatebox[origin=c]{90}{ALL}\\
\cline{2-13}
\textbf{Model} $\downarrow$ \textbf{\#} $\rightarrow$ & 30 & 68 & 34 & 102 & 58 & 98 & 138 & 92 & 66 & 90 & 132 & 908 \\
\rowcolor{bloomcolor}\textsc{bloom-7b} & \textbf{0.633} & 0.515 & 0.412 & 0.529 & 0.655 & 0.480 & 0.543 & 0.522 & 0.470 & 0.489 & 0.508 & 0.520 \\
\rowcolor{bloomcolor}\textsc{bloom-176b} & 0.600 & \textbf{0.676} & 0.735 & 0.608 & 0.638 & 0.561 & 0.565 & 0.554 & 0.606 & 0.644 & 0.758 & 0.628\\
\rowcolor{falconcolor}\textsc{falcon-7b} & 0.567 & 0.529 & \textbf{0.500} & 0.559 & 0.517 & 0.541 & 0.587 & 0.500 & 0.561 & 0.556 & 0.644 & 0.561\\
\rowcolor{falconcolor}\textsc{falcon-40b} & 0.800 & 0.882 & 0.882 &  \textbf{0.882} & 0.931 & 0.847 & 0.891 & 0.880 & 0.894 & 0.889 & 0.932 & 0.882\\
\rowcolor{llamacolor}\textsc{llama2-7b} & 0.800 & 0.882 & 0.794 & 0.931 & \textbf{0.948} & 0.898 & 0.826 & 0.891 & 0.818 & 0.811 & 0.902 & 0.871\\
\rowcolor{llamacolor}\textsc{llama2-13b} & 0.867 & 0.912 & 0.853 & 0.951 & 0.966 & \textbf{0.918} & 0.891 & 0.913 & 0.864 & 0.867 & 0.962 & 0.913 \\
\rowcolor{llamacolor}\textsc{llama2-70b} & 1.000 & 1.000 & 1.000 & 1.000 & 1.000 & 0.969 & \textbf{0.986} & 0.967 & 0.955 & 0.989 & 0.992 & 0.986\\
\rowcolor{mistralcolor}\textsc{mistral-7b} & 0.967 & 0.985 & 0.941 & 0.980 & 1.000 & 0.929 & 0.935 & \textbf{0.967} & 0.879 & 0.889 & 0.977 & 0.949\\
\rowcolor{mptcolor}\textsc{mpt-7b} & 0.667 & 0.838 & 0.706 & 0.804 & 0.776 & 0.714 & 0.841 & 0.815 & \textbf{0.591} & 0.656 & 0.811 & 0.764\\
\rowcolor{mptcolor}\textsc{mpt-30b} & 0.800 & 0.882 & 0.882 & 0.941 & 0.966 & 0.796 & 0.848 & 0.880 & 0.894 & \textbf{0.833} & 0.947 & 0.882\\
\rowcolor{phicolor}\textsc{phi-2} & 0.800 & 0.956 & 0.941 & 0.971 & 0.983 & 0.969 & 0.949 & 0.924 & 0.909 & 0.900 & \textbf{0.992} & 0.947\\
\hline
\end{tabular}
\caption{Answering accuracy on valid Gen-COPA items. Each LLM's consistency on its own items is in bold.}
\label{gen_copa_accuracy}
\end{table*}

The consistency results in Section \ref{consistency} are affected by both the LLM's answering performance as well as the quality of its Gen-COPA items. For instance, \textsc{bloom-7b} has the lowest consistency but also the lowest Gen-COPA validity. Its poor performance in answering its own items might be due to how often the \texttt{mpa} is invalid. We considered whether the results would differ if we isolated only the valid Gen-COPA items and measured the LLMs' answering accuracy specifically on these sets. 

Given the valid schemas for a single LLM, we used the same process described in Section \ref{consistency} to randomly map the \texttt{mpa} and \texttt{lpa} in each schema to the numerical labels, balancing them across the set so that always selecting Alternative 1 would yield 50\% accuracy. For this analysis, we ensured an even number of items in each set, so we downsampled one item in sets with an odd number of valid Gen-COPA items. We then applied the same 4-shot prompt format to these items that we used for the experiments in Section \ref{original_copa} and Section \ref{consistency}.

Table \ref{gen_copa_accuracy} shows each LLM's answering accuracy on all valid Gen-COPA sets. The bolded numbers on the diagonal indicate each model's consistency on its own items. For all eleven Gen-COPA sets, the accuracies obtained by the LLMs on the set are strongly correlated with their accuracies on the Orig-COPA test set ($r_s\ge.89, p<.001$ for all columns of Table \ref{gen_copa_accuracy}). This shows that LLMs' answering performance on COPA also generalizes to LLM-generated versions of COPA.  Moreover, the fact that the models that do particularly well on Orig-COPA (i.e. \textsc{llama2-70b}, \textsc{mistral-7b}, and \textsc{phi-2}) also do well on answering Gen-COPA suggests their success on the former is not just an illusion resulting from rote memorization of the Orig-COPA test set. Since we verified the Gen-COPA items are not duplicates of Orig-COPA items, we know the answers to the Gen-COPA items have not been memorized by the LLMs during training.


Accuracy on the valid Gen-COPA sets tends to be higher for all LLMs compared with their accuracy on Orig-COPA. In particular, examining the final column in Table \ref{gen_copa_accuracy} where the results are aggregated over all Gen-COPA sets, the accuracy of each LLM is between 1 and $\approx$11 percentage points higher on Gen-COPA than on Orig-COPA, with exception to \textsc{bloom-7b} whose accuracy is low on both groups of items. This suggests that the valid subset of Gen-COPA items is easier to answer than Orig-COPA. This may reflect the mechanism behind model collapse, where synthetically generated data contains only the most common patterns in the original data distribution, losing the diverse signal in the distribution tail \citep{shumailov2024ai}.

These results also confirm that LLMs do not necessarily answer their own Gen-COPA items consistently even if the items are valid. For example, \textsc{falcon-7b} performs exactly at the level of random guessing on its own valid Gen-COPA set. This finding aligns with increasing evidence demonstrating LLM inconsistency in question answering performance: in particular, changing the order and/or labels of answers can drastically impact the model's accuracy \citep{wang2024answers,zheng2024large}. \citet{wang-etal-2024-answer-c} found that consistency across different answer label assignments increased with model capability. Similarly, an LLM's consistency in answering its own generated questions may be a broad indicator of its abilities; this warrants further analysis with other tasks beyond COPA.

\subsection{Composition Quality}

Just like Orig-COPA items, valid Gen-COPA items each have an agreed-upon alternative that is considered the correct answer. However, even items with a correct answer may not meet all composition quality standards reflected in the original benchmark. As with most benchmarks, while there are some explicit authoring guidelines for COPA, many of the quality standards are only implicitly defined. Thus, we enlisted one of the authors of Orig-COPA items to assess the intrinsic authoring quality of the valid Gen-COPA items. A item rated as \textbf{high-quality} is one the rater deems they would accept as an additional item in the benchmark. 

Table \ref{gen_copa_high_quality} shows the proportion of items categorized as high-quality among the set of valid Gen-COPA items produced by each LLM. Though the association is more moderate than that of validity, the rate of high-quality items is also correlated with answering accuracy on Orig-COPA ($r_s=.76, p=.007$). This again suggests that LLMs that correctly answer the original benchmark also have better authoring ability.

\begin{table}[h]
\small
\centering
\begin{tabular}{ c  c  c }
\textbf{Model} & \textbf{High-Quality}\\
\hline
\rowcolor{bloomcolor}\textsc{bloom-7b} &  0.533\\
\rowcolor{bloomcolor}\textsc{bloom-176b} & 0.500\\
\rowcolor{falconcolor}\textsc{falcon-7b} & 0.500\\
\rowcolor{falconcolor}\textsc{falcon-40b} & 0.569\\
\rowcolor{llamacolor}\textsc{llama2-7b} & 0.448\\
\rowcolor{llamacolor}\textsc{llama2-13b} & 0.541\\
\rowcolor{llamacolor}\textsc{llama2-70b} & 0.616\\
\rowcolor{mistralcolor}\textsc{mistral-7b} & 0.663\\
\rowcolor{mptcolor}\textsc{mpt-7b} & 0.470\\
\rowcolor{mptcolor}\textsc{mpt-30b} & 0.567\\
\rowcolor{phicolor}\textsc{phi-2} & 0.629\\
\hline
\end{tabular}
\caption{Rate of high-quality valid Gen-COPA items}
\label{gen_copa_high_quality}
\end{table}

\begin{table*}[h!]
\small
\centering
\rowcolors{1}{white}{gray!20}
\begin{tabularx}{\textwidth}{ p{0.265\textwidth}  X }
\textbf{Description} & \textbf{Example}\\
\hline

The \texttt{lpa} is not plausible in any context regardless of the premise &
\textbf{Premise}: The woman was arrested.
What was the \textit{cause} of this? [\textsc{falcon-40b}]\newline
\textbf{\texttt{mpa}}: She stole something.\newline
\textbf{\texttt{lpa}}: She was a zombie.\\

The relation between the \texttt{mpa} and the premise is clear but very trivial &
\textbf{Premise}: The students cleaned the beach.
What happened as a \textit{result}? [\textsc{llama2-70b}]\newline
\textbf{\texttt{mpa}}: The beach was clean.\newline
\textbf{\texttt{lpa}}: The beach was polluted.\\

The \texttt{mpa} is more plausible than the \texttt{lpa}, but the relation is more temporal or semantic than causal &
\textbf{Premise}: The sales associate spoke to the customer.
What was the \textit{cause} of this? [\textsc{phi-2}]\newline
\textbf{\texttt{mpa}}: The customer made a purchase.\newline
\textbf{\texttt{lpa}}: The sales associate went on vacation.\\

The \texttt{lpa} is also a plausible causal relation, but the \texttt{mpa} is a closer temporal relation to the premise&
\textbf{Premise}: I broke my nose.
What happened as a \textit{result}? [\textsc{bloom-7b}]\newline
\textbf{\texttt{mpa}}: My nose started bleeding.\newline
\textbf{\texttt{lpa}}: I took a week off work.\\

The relation between the \texttt{mpa} and premise is clear only with assumptions not defined in the premise &
\textbf{Premise}: The landlord is sending eviction notices.
What happened as a \textit{result}? [\textsc{phi-2}]\newline
\textbf{\texttt{mpa}}: The tenant had to leave.\newline
\textbf{\texttt{lpa}}: The landlord offered a lower rent.\\
\hline
\end{tabularx}
\caption{Common quality problems in valid Gen-COPA items}
\label{weak_valid_patterns}
\end{table*}

To illuminate the characteristics underlying these ratings, Table \ref{weak_valid_patterns} describes and exemplifies the most common reasons that an item failed to receive a high-quality rating, while Table \ref{successful_examples} lists examples of items that were marked as high-quality. Notably, in these high-quality items, the \texttt{lpa} is highly related to the premise, but its causal relation is unclear compared with the \texttt{mpa}. Consequently, these items are challenging to answer without discerning causal relatedness separately from coarse semantic relatedness. This is a prominent feature of Orig-COPA items and was a key reason the benchmark remained unbeaten for so long. All Gen-COPA items with their validity and quality annotations are publicly available for further analysis\footnote{\href{https://huggingface.co/datasets/roemmele/Gen-COPA}{huggingface.co/datasets/roemmele/Gen-COPA}}.

\begin{table}[h!]
\small
\centering
\rowcolors{1}{white}{gray!20}
\begin{tabularx}{\columnwidth}{ X }
\hline
\textbf{Premise}: The girl gave her friend her lunch.\newline
What was the \textit{cause} of this? [\textsc{mpt-30b}]\newline
\textbf{\texttt{mpa}}: She was concerned about her friend's lack of food.\newline
\textbf{\texttt{lpa}}: Her friend sat next to her at lunch.\\

\textbf{Premise}: The student x-rayed the patient's arm.\newline
What happened as a \textit{result}? [\textsc{mpt-30b}]\newline
\textbf{\texttt{mpa}}: He discovered that the patient's arm was broken.\newline
\textbf{\texttt{lpa}}: He put the patient's arm in a cast.\\

\textbf{Premise}: The movie star was in seclusion.\newline
What was the \textit{cause} of this? [\textsc{falcon-40b}]\newline
\textbf{\texttt{mpa}}: The movie star needed a break from the spotlight.\newline
\textbf{\texttt{lpa}}: The movie star found a secret underground club.\\

\textbf{Premise}: The reader was puzzled by a joke.\newline
What happened as a \textit{result}? [\textsc{falcon-40b}]\newline
\textbf{\texttt{mpa}}: She looked up the explanation.\newline
\textbf{\texttt{lpa}}: She laughed at the joke.\\

\textbf{Premise}: The woman destroyed the artwork.\newline
What was the \textit{cause} of this? [\textsc{llama2-70b}]\newline
\textbf{\texttt{mpa}}: The artwork was an insult to her religion.\newline
\textbf{\texttt{lpa}}: The artwork was an advertisement for a store.\\
\hline
\end{tabularx}
\caption{Examples of high-quality Gen-COPA items}
\label{successful_examples}
\end{table}





\section{Related Work and Outlook}

Researchers are increasingly turning to LLMs to replace human effort in developing and evaluating NLP systems. In particular, LLMs are being used to synthesize labeled data in order to train or fine-tune models \citep{choi-etal-2024-gpts,he2022generate,li-etal-2023-synthetic}. LLMs are also being applied to score the output quality of other models \citep{chiang-lee-2023-large,kocmi-federmann-2023-large,wang2023chatgpt}. Our work aligns with the above endeavors in exploring the use of LLMs as an alternative to human authoring of evaluation items. 

Our particular goal in using LLMs for this assessment authoring is to better understand their capabilities, rather than to produce a novel benchmark that is valuable for evaluating future LLMs. However, there are some emerging demonstrations of using LLMs to facilitate the creation of evaluation data that is unique in its design and scope. For instance, \citet{anthropic2023} described prompting an LLM to derive multiple-choice Q\&A sets from short sections of text, which are then used to evaluate the same LLM's ability to answer those questions when the relevant information is provided in one long document. As another example, \citet{tian-etal-2024-macgyver} utilized an LLM as an interactive tool combined with human feedback in the authoring process for a novel challenge set focused on physical problem-solving.

Given the short lifecycle of benchmarks, rapid creation of new ones is pivotal to capturing further progress in NLP. Additionally, because they do not serve a broader purpose outside of research, the authoring standards for benchmarks are not always clearly understood by the human authors tasked with their creation, and these standards are not necessarily well-anchored to other real-world writing tasks. The current paradigm of LLM interaction via few-shot prompting suggests that LLMs can perform sophisticated authoring tasks in the absence of explicit guidelines just by observing representative examples. As current benchmarks quickly age and new benchmarks become more complex in what they aim to measure, LLMs are likely to take on a sanctioned role in their development.


\section{Conclusion}

This paper looks at a notable assessment of commonsense reasoning, COPA, as an authoring task performed by LLMs. The results indicate that models that answer COPA items correctly (both LLM-authored and human-authored ones) are also better at writing COPA items. In future work, we will investigate how this extends to other benchmarks. 

Our work bears upon the trend of widening generalizability of NLP models across tasks. Previously, models designed for text generation were not considered directly applicable to commonsense question answering benchmarks, because this answering task was presumed to require a distinct model architecture supporting answer label prediction. This is no longer a constraint in the new paradigm of LLMs. Increasingly, we are observing models demonstrate the same core knowledge across highly diverse task representations.

\section*{Limitations}

The current paradigm of LLMs has some key limitations that are reflected in this work. First, there has been limited transparency into the development process behind most LLMs, even the open-weight models used here. With exception to \textsc{bloom} which had transparency as an explicit goal of its development, not all details of the LLMs in this work regarding their training data, model architecture, and optimization techniques have been clearly documented. Given this as well as the complexity of these details, it is not easy to interpret why exactly different types of LLMs perform differently on the same benchmarks. So while we conclude there is a positive association between answering and generating COPA, we do not propose an explanation for why certain models (e.g. the \textsc{llama2} family) excel on these tasks more than other others (e.g. the \textsc{bloom} family). 

Second, LLM behavior is highly sensitive to prompt design, such that different prompts representing the same task and input features can yield different outputs. As a result, prompt optimization has become a significant focus of utilizing LLMs. Finding tractable solutions to this optimization process is an ongoing research endeavor \citep[e.g.][]{zhou2023large}. Meanwhile, this is typically done manually without any guarantee of optimality. In particular, researchers often write a few different prompt variations based on their knowledge of the task and select the one that performs best on the task evaluation. In this work, we took a principled rather than data-driven approach to selecting the prompts for answering and generating COPA, in which the prompt design matches the conceptual design of the benchmark defined in \citet{roemmele2011choice}. It is possible that varying the language of the prompt as well as the number of exemplars (shots) would result in better performance in terms of answering accuracy or Gen-COPA validity for some LLMs, which could yield a different view of how they compare to one another. 

\section*{Ethical Considerations}

The business of data annotation has grown dramatically with the expansion of NLP systems and LLMs in particular. There are serious concerns about the ethics of this business when it comes to fair compensation of workers \citep[e.g.][]{perrigo2023}. For the annotation in this work, the expert rater was compensated as part of their normal job role. The external raters we employed on Prolific were paid at the rate of \$12/hour, which meets Prolific’s recommended universal standard for fair pay \citep{prolific-payment}.

LLMs have a well-known risk of generating offensive content \cite[e.g.][]{gehman-etal-2020-realtoxicityprompts}. We anticipated this risk would emerge in some of the Gen-COPA items and sought to reduce exposure to these items to people outside our internal team. As described in Section \ref{gen_copa_validity_section}, the expert rater consented to conduct an initial review of all Gen-COPA items. They assigned a “content warning” to items they deemed potentially offensive or harmful, and these items were consequently not shown to external raters on Prolific. While we cannot guarantee that the external raters were not offended by items that were not flagged in the initial review, we did not receive any report of this from raters.

\section*{Acknowledgements}

(Gordon) The project or effort depicted was or is sponsored by the U.S. Army Research Laboratory (ARL) under contract number W911NF-14-D-0005, and that the content of the information does not necessarily reflect the position or the policy of the Government, and no official endorsement should be inferred.

\bibliography{anthology,custom}

\appendix
\section{Novelty of Gen-COPA with regard to Orig-COPA}\label{gen_copa_novelty_section}

As described in Section \ref{llm_authoring_gen_copa}, we verified that none of the Gen-COPA items used in our analyses are exact duplicates of Orig-COPA items. However, it's important to consider the overall similarity between Gen-COPA and Orig-COPA, to ascertain that the Gen-COPA items are not just trivial variations of Orig-COPA items. Table \ref{gen_copa_novelty} reports two metrics pertaining to this for the passable Gen-COPA sets produced by each LLM. The first metric, \textbf{Common 3-grams}, indicates the overall proportion of trigrams in the Gen-COPA items that also appear in Orig-COPA. These percentages range from $\approx$8\% to $\approx$12\% for all LLMs, indicating that the majority of trigrams in the Gen-COPA items are not contained in Orig-COPA. The second metric pertains to ROUGE-3 F1, which is computed to determine the similarity of each Gen-COPA item to each of the Orig-COPA items. For a given Gen-COPA item, the maximum of these scores is selected, which corresponds to the Orig-COPA item that is most similar to that Gen-COPA item in terms of trigram overlap. \textbf{Max ROUGE-3} reports the mean of these maximum scores across the Gen-COPA items for each LLM. These scores range from 0.058 to 0.091 for all LLMs, indicating that on average a particular Gen-COPA item has only a minimal degree of trigram overlap with the Orig-COPA item it most closely resembles. Thus, the majority of Gen-COPA items are reasonably distinct from Orig-COPA items.

\begin{table}[h!]
\small
\centering
\begin{tabular}{ c  c  c }
\textbf{Model} & \textbf{Common 3-grams} & \textbf{Max ROUGE-3}\\
\hline
\rowcolor{bloomcolor}\textsc{bloom-7b} & 0.079 & 0.058 \\
\rowcolor{bloomcolor}\textsc{bloom-176b} & 0.079 & 0.063\\
\rowcolor{falconcolor}\textsc{falcon-7b} & 0.091 & 0.067\\
\rowcolor{falconcolor}\textsc{falcon-40b} & 0.090 & 0.070\\
\rowcolor{llamacolor}\textsc{llama2-7b} & 0.089 & 0.063\\
\rowcolor{llamacolor}\textsc{llama2-13b} & 0.093 & 0.063\\
\rowcolor{llamacolor}\textsc{llama2-70b} & 0.096 & 0.067\\
\rowcolor{mistralcolor}\textsc{mistral-7b} & 0.085 & 0.059\\
\rowcolor{mptcolor}\textsc{mpt-7b} & 0.091 & 0.065\\
\rowcolor{mptcolor}\textsc{mpt-30b} & 0.123 & 0.091\\
\rowcolor{phicolor}\textsc{phi-2} & 0.096 & 0.071\\
\hline
\end{tabular}
\caption{Redundancy of Gen-COPA with Orig-COPA}
\label{gen_copa_novelty}
\end{table}

\end{document}